\documentclass[runningheads]{llncs}
\usepackage{graphicx}
\usepackage{booktabs,caption}
\pdfoutput=1
\usepackage[flushleft]{threeparttable}
\usepackage{amsmath,amssymb} 
\usepackage{mathtools}
\usepackage{color}
\usepackage[width=122mm,left=12mm,paperwidth=146mm,height=193mm,top=12mm,paperheight=217mm]{geometry}
\usepackage{multirow}
\usepackage[table,xcdraw]{xcolor}
\usepackage{subcaption}
\usepackage{hhline}
\usepackage{authblk}
\DeclarePairedDelimiter\abs{\lvert}{\rvert}%
\DeclarePairedDelimiter\norm{\lVert}{\rVert}%
\let\oldabs\abs
\def\abs{\@ifstar{\oldabs}{\oldabs*}}

\newcommand{\squeezeup}{\vspace{-5.5mm}}
\makeatletter
\let\oldnorm\norm
\def\norm{\@ifstar{\oldnorm}{\oldnorm*}}
\makeatother

\mainmatter
\begin{document}

\def\ECCV18SubNumber{xxxx}  

\title{Fast View Synthesis with Deep Stereo Vision} 

\titlerunning{Fast View Synthesis with Deep Stereo Vision}

\authorrunning{Habtegebrial et. al.}

\author{
Tewodros Habtegebrial\inst{1, 2}  \and 
Kiran Varanasi\inst{2}  \and 
Christian Bailer\inst{2}  \and 
Didier Stricker \inst{1, 2}
}
\institute{
TU Kaiserslautern
\and
German Research Center for Artificial Intelligence}


\maketitle

\begin{abstract}

Novel view synthesis is an important problem in computer vision and graphics. Over the years a large number of solutions have been put forward to solve the problem. However, the large-baseline novel view synthesis problem is far from being "solved". Recent works have attempted to use Convolutional Neural Networks~(CNNs) to solve view synthesis tasks. Due to the difficulty of learning scene geometry and interpreting camera motion, CNNs are often unable to generate realistic novel views. In this paper, we present a novel view synthesis approach based on stereo-vision and CNNs that decomposes the problem into two sub-tasks: \textit{view dependent geometry} estimation and \textit{texture inpainting}. Both tasks are structured prediction problems that could be effectively learned with CNNs. Experiments on the KITTI Odometry dataset show that our approach is more accurate and significantly faster than the current state-of-the-art. The code and supplementary material will be publicly available\footnote{Results could be found here \url{https://youtu.be/5pzS9jc-5t0}}.

\end{abstract}

\begin{figure}[htb!]
\centering
\includegraphics[width=\textwidth]{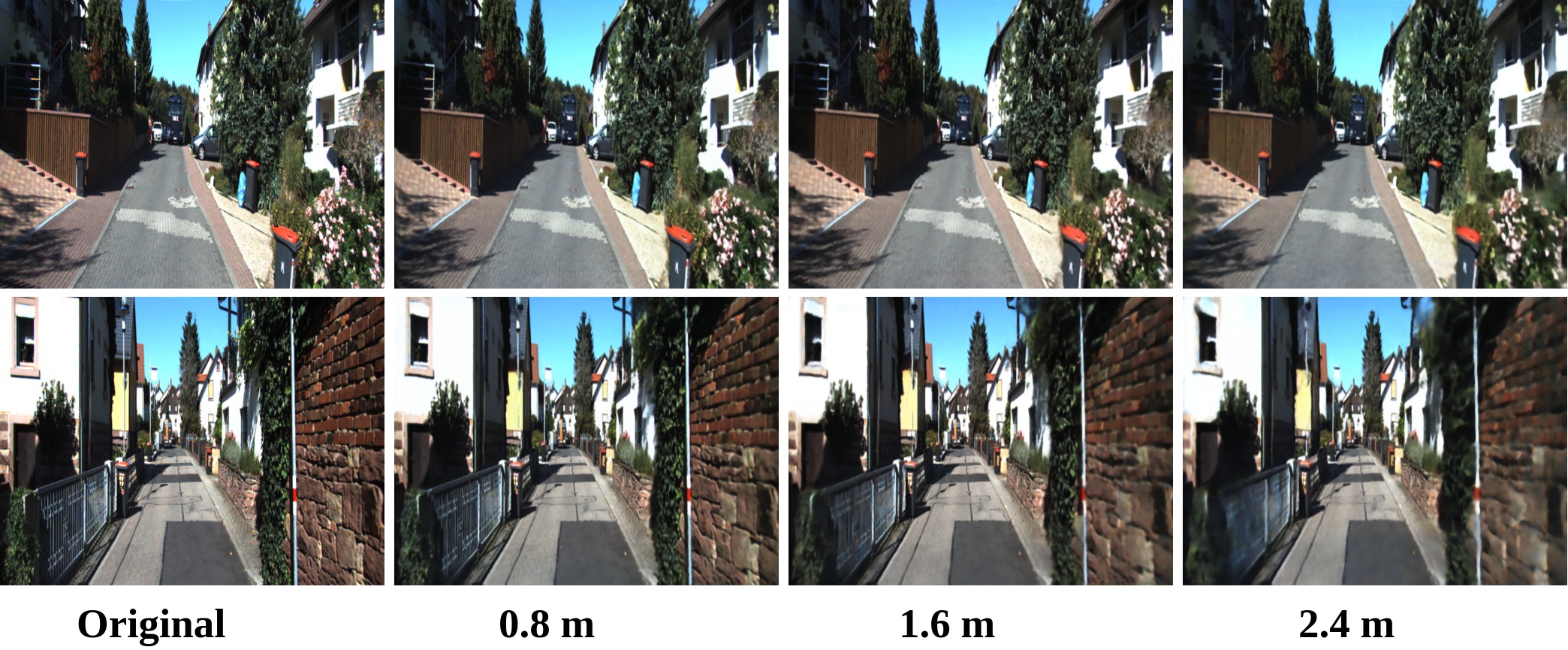}
\caption{Two sample renderings from the KITTI dataset~\cite{geiger2012we}, using our proposed method. Images are rendered using four neigbouring views. From left to right, the median baseline between consecutive input views increases from $0.8m$ to $2.4m$.}
\label{fig:qualitative}
\end{figure}
\squeezeup
\section{Introduction}

Novel view synthesis~(NVS) is defined as the problem of rendering a scene from a previously unseen camera viewpoint, given other reference images of the same scene. This is an inherently ambiguous problem due to perspective projection, occlusions in the scene, and the effects of lighting and shadows that vary with a viewpoint. Due to this inherent ambiguity, this can be solved only by learning valid scene priors and hence, is an effective problem to showcase the application of machine learning to computer vision. Accurate rendering of novel views can be a useful component for many computer vision applications, such as pedestrian detection or robot tool positioning, whose accuracy is significantly affected by the viewpoint. Many applications require extremely robust and fast or even real-time NVS, which is currently beyond the state of the art. Furthermore, NVS is essential for rendering real-world scenes in virtual reality.

In the early 1990s, methods for NVS were proposed to deal with slight viewpoint changes, given images taken from relatively close viewpoints. Then NVS can be performed through view interpolation~\cite{chen_and_wiiliams}, warping~\cite{seitz} or rendering with stereo reconstruction~\cite{scharstein1996stereo}. A dense matrix of camera views can be used to sample the \textit{plenoptic function} ~\cite{adelson1991plenoptic}, which can be rendered into new viewpoints. Levoy and Hanrahan~\cite{levoy1996light}, McMillan and Bishop~\cite{mcmillan1995plenoptic} introduced image-based rendering (IBR) as an attempt to reconstruct novel views from these samples of plenoptic function. This framework is not very suitable when the plenoptic function is sampled sparsely i.e when the input camera views are separated by a large baseline. Across wide baselines, NVS is a particularly challenging problem due to sharp foreshortening effects, scale changes and the 3D rotation of objects. In this paper, we address this challenging problem and demonstrate results on the KITTI dataset, where stereo pairs of images were recorded from a moving car.
\newline \\
There are a few methods proposed over the years to solve the problem of large-baseline NVS~\cite{chaurasia2013depth}, \cite{zitnick2004high}, \cite{flynn2015deepstereo}, \cite{goesele2010ambient}, \cite{penner2017soft}. Some methods are based on structure from motion~(\textit{SFM})~\cite{zitnick2004high}, which can produce high-quality novel views in real time~\cite{chaurasia2013depth}, but has limitations when the input images contain strong noise, illumination change and highly non-planar structures like vegetation. These methods need to produce depth synthesis for poorly reconstructed areas in \textit{SFM}, which is challenging for intricate structures. In contrast to these methods, neural networks can be trained end-to-end to render NVS directly~\cite{flynn2015deepstereo}. This is the paradigm we follow in this paper.

Many recent works have addressed different facets of end-to-end training of neural networks for NVS ~\cite{dosovitskiy2015learning}, \cite{tatarchenko2016multi}, \cite{zhou2016view}, \cite{kulkarni2015deep}, \cite{yang2015weakly}. These methods typically perform well under a restricted scenario, where they have to render geometrically simple scenes, like a single object rendered from the ShapeNet Dataset ~\cite{shapenet}, but addressing real-world scenes with large variations is still challenging. The state-of-the-art large-baseline view synthesis approach that works well under challenging scenarios is \textit{DeepStereo}, 
introduced by Flynn et. al.~\cite{flynn2015deepstereo}. \textit{DeepStereo} generates high-quality novel views on the KITTI dataset~\cite{geiger2012we}, where older \textit{SFM} based methods such as~\cite{chaurasia2013depth} do not work at all. This algorithm uses plane-sweep volumes and processes them with a double tower CNN (with \textit{color} and \textit{selection} towers). However, processing plane-sweep volumes of all reference views jointly imposes very high memory and computational costs. Thus, \textit{DeepStereo} is far slower than previous methods \textit{SFM} based methods such as~\cite{chaurasia2013depth}.

In this work we propose a novel alternative to \textit{DeepStereo}, which is two orders of magnitude faster. Our method avoids performing expensive calculations on the combination of plane-sweep volumes of reference images. Instead, we predict a proxy scene geometry for the input views with stereo-vision. Using forward-mapping (section~\ref{sec:fwdmap}), 
we project input views to the novel view. Forward-mapped images contain a large number of pixels with unknown color values. Rendering of the target view is done by applying texture inpainting on the warped-images. Our rendering pipeline is fully-learnable from a sequence of calibrated stereo images. Compared to \textit{DeepStereo}, the proposed approach produces more accurate results while being significantly faster. The main contributions of this paper are the following.

\begin{itemize}
\item{We present a novel view synthesis approach based on stereo-vision. The proposed approach decomposes the problem into \textit{proxy} geometry prediction and texture inpainting tasks. Every part of our method is fully learnable from input stereo reference images.}

\item{Our approach provides an affordable large-baseline view synthesis solution. The proposed method is faster and even more accurate than the current \textit{state-of-the-art} \cite{flynn2015deepstereo} large baseline view synthesis. The proposed approach takes seconds to render a single frame while DeepStereo \cite{flynn2015deepstereo} takes minutes.}
\end{itemize}

\section{Related Work}
Image based rendering has enjoyed a significant amount of attention from the computer vision and graphics communities. Over the last few decades, several approaches of image-based rendering and modelling were introduced~\cite{chen_and_wiiliams}, \cite{seitz}, \cite{seitz2}, \cite{adelson1991plenoptic}, \cite{scharstein1996stereo}. Fitzgibbon et. al.~\cite{fitzgibbon2005image} present a solution that solves view synthesis as texture synthesis by using image-based priors as regularization. Chaurasia et. al.~\cite{chaurasia2013depth}, presented high-quality view synthesis that utilizes 3D reconstruction. Recently, Penner at. al.~\cite{penner2017soft} presented a view synthesis method that uses soft 3D reconstruction via fast local stereo-matching similar to \cite{hosni2011real} and occlusion aware depth-synthesis. Kalantari et. al. \cite{kalantari2016learning} used deep convolutional networks for view synthesis in light-fields.
\newline
\subsubsection*{Encoder-decoder Networks} have been used in generating unseen views of objects with simple geometric structure, (e.g. cars, chairs, etc)~\cite{tatarchenko2016multi}, \cite{dosovitskiy2015learning}. 
However, the renderings generated from encoder-decoder architectures are often blurry. Zhou et al.~\cite{zhou2016view}, used encoder-decoder networks to predict appearance flow, rather than directly generating the image. Compared to direct novel view generating methods~\cite{tatarchenko2016multi}, \cite{dosovitskiy2015learning}, the appearance-flow based method produces crispier results. Nonetheless, the appearance flow based also fails to produce any convincing results in natural scenes.
\newline
\subsubsection*{Deep Generative Models} As it has already been done in encoder-decoder networks~\cite{dosovitskiy2015learning}, \cite{tatarchenko2016multi}, view synthesis could be proposed as a generative modeling task. A generative model of view synthesis would have to hallucinate what a scene looks like from a certain camera pose, given reference views of the scene. Most of the recent generative neural networks such as ~\cite{isola}, \cite{pix2pixhd}, \cite{chen2017photographic}, however, work in scenarios where almost no strong geometric manipulation of the input is needed. For instance, converting images to semantic labels and vice versa. This limits their applicability in novel view synthesis.
\newline
\subsubsection*{DeepStereo} Flynn et. al. ~\cite{flynn2015deepstereo}, proposed the first CNN based large baseline novel view synthesis approach. DeepStereo has a double-tower(\textit{color \text{and} selection towers}) CNN architecture. DeepStereo takes a volume of images projected using multiple depth planes, known as \textit{plane-sweep volume} as input. The \textit{color-tower} produces renderings for every depth plane separately. The \textit{selection-tower} estimates probabilities for the renderings computed for every depth plane. The output image is then computed as a weighted average of the rendered color-images. DeepStereo generates high-quality novel views from a plane sweep volume generated from few(typically 4) reference views. To the best of our knowledge, \textit{DeepStereo} is the most accurate large-baseline method, proven to be able to generate accurate novel views of challenging natural scenes.
\newline
\subsubsection*{Depth Prediction based view synthesis approaches} Supervised depth prediction with CNNs have been widely studied by the computer vision community~\cite{eigen}, \cite{zbontar2016stereo}, \cite{laina2016deeper}.  Recent works~\cite{godard},\cite{zhou2017} demonstrated that CNN based monocular depth prediction could be learned even in the absence of ground truth depth. Godard et. al.~\cite{godard} and Zhou et. al.~\cite{zhou2017}, learn to predict depth by using stereo-reconstruction~\cite{godard} and multi-view reconstruction losses, respectively. In addition to monocular depth prediction, CNN based methods have been successful in stereo depth prediction tasks~\cite{gcnet}. Kendall et. al.~\cite{gcnet} presented a fast and accurate supervised stereo-depth prediction network called Geometry and Context Network(GCNet). Recently, monocular depth prediction based view synthesis methods have been proposed ~\cite{mono1}, \cite{mono2}. Despite being fast, these works produce results with significantly lower quality compared to multiview approaches such as \textit{DeepStereo}~\cite{flynn2015deepstereo}.
\newline


\begin{figure}[htb!]
 
\begin{minipage}[b]{\linewidth}
  \centering
  \includegraphics[width=\linewidth]{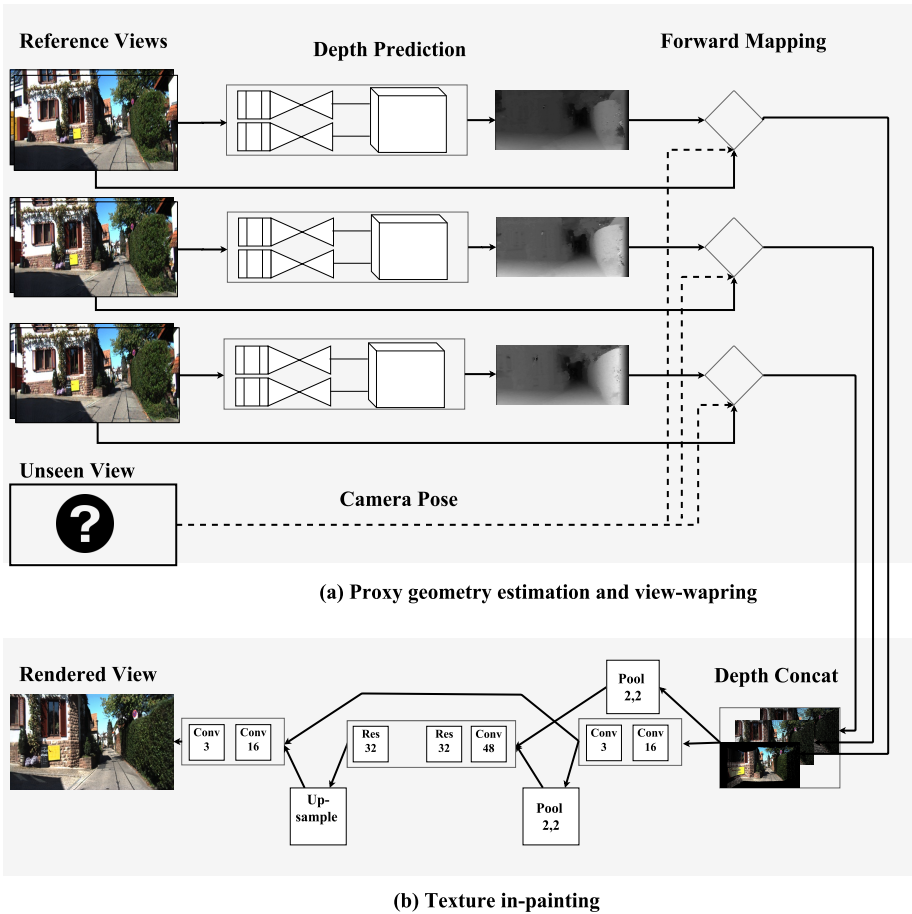}
  \caption{Illustration of our novel view synthesis approach. in the first stage(a), we  begin by estimating dense depth maps from the input reference stereo pairs using an unsupervised stereo-depth prediction network. Estimated depth maps are used to project input views to the target view via Forward-mapping, shown in Equation~\ref{eqn:one}. As shown in (b), the output novel view is rendered with the \textit{texture inpainting} applied on the forward mapped views.}
  \label{fig:system}
\end{minipage}

\end{figure}

\section{Proposed Method}

\sloppy In this section we discuss our proposed view synthesis approach. Our aim is to generate the image $\mathcal{X}^{t}$ of a scene from a novel viewpoint, using a set of reference stereo-pairs $\{\{\mathcal{X}^{1}_{L}, X^{1}_{R}\},\{\mathcal{X}^{2}_{L}, \mathcal{X}^{2}_{R}\}, \dots,  \{\mathcal{X}^{V}_{L}, \mathcal{X}^{V}_{R}\}\}$\footnote{subscripts L and R indicate left and right views, respectively} and their poses $\{\mathcal{P}^{1}, \mathcal{P}^{2}, \dots, \mathcal{P}^{V}\}$ w.r.t the target view. The proposed method has three main stages, namely \textit{proxy scene geometry estimation}, \textit{forward-mapping} and \textit{texture inpainting}. As shown in Figure~\ref{fig:system} the view synthesis is performed as follows: first, a proxy scene geometry is estimated as a dense depth map. The estimated depth map is used to forward-map the input views to the desired novel viewpoint. Forward mapping (described in section~\ref{sec:fwdmap}) leads to noisy images with a large number of holes. The final rendering is, therefore, generated by applying \textit{texture inpainting} on the forward-mapped images. Both the depth estimation and texture inpainting tasks are learned via convolutional networks. Components of our view synthesis pipeline are trainable using a dataset that contains a sequence of stereo-pairs with known camera poses. 
\newline

\subsection{Depth Prediction Network}
\begin{figure}[htb!]
  \centering
  \includegraphics[width=\linewidth]{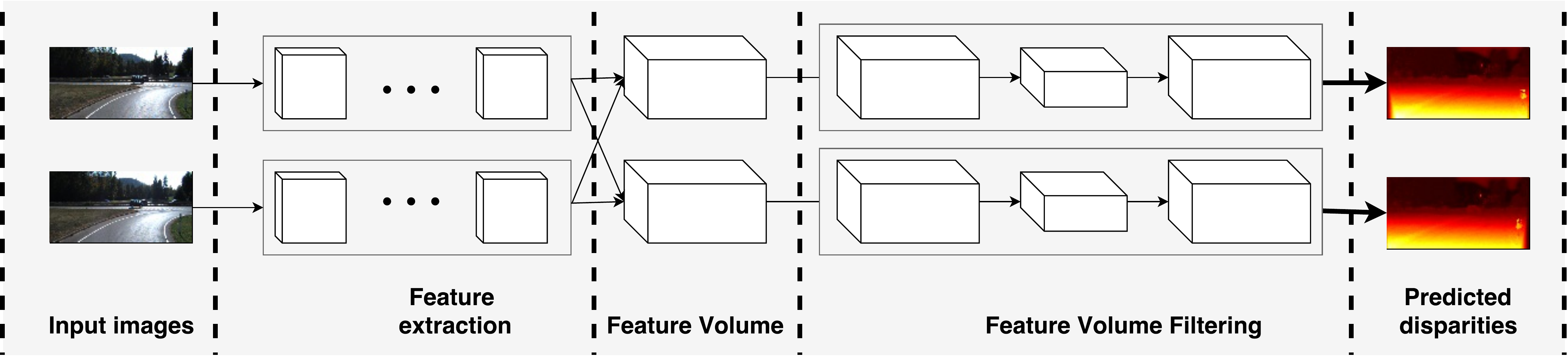}
  \caption{Our depth prediction network architecture.}
  \label{fig:depth}
\end{figure}

We train a convolutional network to estimate depth from an input stereo pair. The training set for the depth prediction CNN is generated by sampling $M$ stereo pairs from our training set. The proposed network shown in Figure~\ref{fig:depth}, is composed of the following stages: \textit{feature extraction}, \textit{feature-volume aggregation} and \textit{feature-volume filtering}. Architecture-wise our network is similar to  GCNet~\cite{gcnet}. However, there are several key differences, including the fact that our network is trained in an unsupervised manner. Similar to our depth prediction network, a recent work \cite{zhong2017self} also investigates learning stereo disparity prediction without using ground truth data.
\newline
\subsubsection*{Feature Extraction:} Generating robust feature descriptors is an important part of stereo matching and optical flow estimation systems. Recently, CNN based feature generation and matching has been used in stereo matching and optical flow estimation papers~\cite{bailer_cvpr}, \cite{gadot2016patchbatch}, \cite{bailer_bmvc}, \cite{zbontar2016stereo}. In this work we extract image-features using a convolutional network. Applying the fully convolutional network~(Table~\ref{table:feat_extractor}) on \textit{left} and \textit{right} stereo-images, we extract features ${\mathcal{F}_{L}}$ and ${\mathcal{F}_{R}}$.
\begin{table}[htb!]
\centering
\begin{tabular}{|l|l|l|l|l|l|}
\hline
Layers      & \begin{tabular}[c]{@{}l@{}}Kernel\\ size\end{tabular} & Stride & \begin{tabular}[c]{@{}l@{}}Input \\ channels\end{tabular} & \begin{tabular}[c]{@{}l@{}}Output\\ channels\end{tabular} & Nonlinearity \\ \hline
conv\_0         & 5x5                                                   & 2x2    & 32                                                        & 32                                                        & ReLU         \\ \hline
res\_1 to res\_9 & 3x3                                                   & 1x1    & 32                                                        & 32                                                        & ReLU + BN    \\ \hline
conv\_10        & 3x3                                                   & 1x1    & 32                                                        & 32                                                        & -            \\ \hline
\end{tabular}
\caption{\textbf{Details of the feature extraction stage.} In the first entry of the third row, we use \textit{res\_1 to res\_9}, as a shorthand notation for a stack of 9 identical residual layers.}
\label{table:feat_extractor}
\end{table}
\squeezeup
\squeezeup

\subsubsection*{Feature-volume Generation:} \sloppy Features ${\mathcal{F}_{L}}$ and ${\mathcal{F}_{R}}$ are aggregated into feature-volumes $\mathcal{V}_{L} \text{ and } \mathcal{V}_{R}$ for the left and right images, respectively. Feature volumes are data structures that are convenient for matching features. Feature volume $\mathcal{V}_{L}$ is created by concatenating $\mathcal{F}_{L}$ with $\mathcal{F}_{R}$ translated at different disparity levels $d_{i} \in \{d_{1}, d_{2}, \dots d_{D}\}$. $\mathcal{V}_{L} = [( \mathcal{F}_{L}, \mathcal{F}^{d_{1}}_{R}), (\mathcal{F}_{L}, \mathcal{F}^{d_{2}}_{R}), \dots, (\mathcal{F}_{L}, \mathcal{F}^{d_{D}}_{R})]$\footnote{We used $(X, Y)$ to denote \textit{concatenating} X and Y across the first dimension and we use $[\text{ X and Y}]$ to denote \textit{stacking} $X \text{ and } Y$, which creates a new first dimension} Similarly, $\mathcal{V}_{R}$ is created by translating $\mathcal{F}_{L}$ and concatenating it with $\mathcal{F}_{R}$. 

\subsubsection*{Feature-volume Filtering:} \sloppy Generated feature volumes aggregate left and right image features at different disparity levels. This stage is posed with the task of computing pixel-wise probabilities over disparities from feature volumes. Since $\mathcal{V}_{L}$ and $\mathcal{V}_{R}$ are composed of features vectors spanning 3-dimensions~(disparity, image-height, and image-width) it is convenient to use 3D convolutional layers. 3D convolutional layers are able to utilize neighborhood information across the above three axes. 
The 3D-convolutional encoder-decoder network used in this work is presented in Table 2. All layers, except the last layer~(\textit{tr\_conv3d\_out}) are followed by a ReLU Non-Linearity and Batch Normalization. 

\begin{table}[htb!]
\centering
\begin{tabular}{|c|c|c|c|}
\hline
Layer       & Input                  & Output size               & Stride  \\ \hline
conv3d\_1   & Feature Volume         & {(}D, 32, H/2, W/2{)}   & (1,1,1) \\ \hline
conv3d\_2   & conv3d\_1              & {(}D, 32, H/2, W/2{)}   & (1,1,1) \\ \hline
conv3d\_3   & conv3d\_2              & {(}D, 32, H/4, W/4{)}   & (2,2,2) \\ \hline
conv3d\_4   & conv3d\_3              & {(}D, 32, H/4, W/4{)}   & (1,1,1) \\ \hline
conv3d\_5   & conv3d\_4              & {(}D, 32, H/4, W/4{)}   & (1,1,1) \\ \hline
conv3d\_6   & conv3d\_5              & {(}D, 32, H/8, W/8{)}   & (2,2,2) \\ \hline
conv3d\_7   & conv3d\_6              & {(}D, 32, H/8, W/8{)}   & (1,1,1) \\ \hline
conv3d\_8   & conv3d\_7              & {(}D, 32, H/16, W/16{)} & (2,2,2) \\ \hline
tr\_conv3d\_1 & conv3d\_8              & {(}D, 32, H/8, W/8{)}   & (2,2,2) \\ \hline
tr\_conv3d\_2 & conv3d\_7, tr\_conv3d\_1 & {(}D, 32, H/4, W/4{)}   & (2,2,2) \\ \hline
tr\_conv3d\_3 & conv3d\_5, tr\_conv3d\_2 & {(}D, 32, H/2, W/2{)}   & (2,2,2) \\ \hline
tr\_conv3d\_4 & conv3d\_2,tr\_conv3d\_3  & {(}D, 32, H, W{)}         & (2,2,2) \\ \hline
output      & tr\_conv3d\_4            & {(}D, 1, H, W{)}         & (1,1,1 ) \\ \hline
\end{tabular}
\label{table:encdec}
\caption{\textbf{Details of the 3D convolutional encoder-decoder network used for feature volume filtering.} Layers with "conv3d\_" are 3d convolutional layers while names that start with "tr\_conv3d\_" represent 3d transposed convolution.}
\end{table}

Let's denote the output of applying feature-volume filtering on $\mathcal{V}_{L}$ and $\mathcal{V}_{R}$ as $\mathcal{C}_{L}$ and $\mathcal{C}_{R}$, respectively. $\mathcal{C}_{L}$ and $\mathcal{C}_{R}$, are tensors represent pixel-wise disparity "confidences". In order to convert confidences into pixel-wise disparity maps, a \textit{soft arg-min} function is used. $\mathcal{C}_{L}$ and $\mathcal{C}_{R}$ represent negative of confidence, hence, we use \textit{soft-argmin}, instead of \textit{soft-argmax}. The \textit{soft-argmin} operation~(Equation~\ref{eqn:softmin}) is a differentiable alternative to \textit{argmin}~\cite{gcnet}. For every pixel location, \textit{soft-argmin} first normalizes the depth confidences into a proper probability distribution function. Then, disparities are computed as the expectation of the disparity under the normalized distributions. Thus applying \textit{soft-min} on $\mathcal{C}_{L}$ and $\mathcal{C}_{R}$ gives disparity maps $\mathcal{D}_{L}$ and $\mathcal{D}_{R}$, respectively.

\begin{equation}
    \begin{aligned}
        \mathcal{P}_{L}(i,x,y) &= \frac{e^{-\mathcal{C}_{L}(i,x,y)}}{\sum_{j=1}^{D}{e^{-\mathcal{C}_{L}(j,x,y)}}}
    \vspace{0.1cm} \\
    \mathcal{D}_{L}(x,y) =& \sum_{i=1}^{D}{disp(i)*\mathcal{P}_{L}(i,x,y)}
    \end{aligned}
    \label{eqn:softmin}
\end{equation}

The estimated disparities $\mathcal{D}_{L}, \mathcal{D}_{R}$ are disparities that encode motions of pixels between the left and right images. We convert the predicted disparities into a sampling grid in order to warp left image to the right image~(and vice versa). Once sampling grid is created we apply bi-linear sampling \cite{jaderberg2015spatial} to perform the warping. Denoting the bi-linear sampling operation as $\Phi$, the warped left and right images could be expressed as $\Tilde{\mathcal{X}}_{L} = \Phi(\mathcal{X}_{R}, \mathcal{D}_{L})$ and $\Tilde{\mathcal{X}}_{R} = \Phi(\mathcal{X}_{L}, \mathcal{D}_{R})$, respectively.

\subsubsection*{Disparity Estimation Network Training Objective}
Our depth prediction network is trained in unsupervised manner by minimizing a loss term which mianly depends on the photometric discrepancy between the input images $\{\mathcal{X}_{L} \text{,} \mathcal{X}_{R}\}$ and their respective re-renderings $\{\Tilde{\mathcal{X}}_{L} \text{,} \Tilde{\mathcal{X}}_{R}\}$. Our network minimizes, the loss term $\mathcal{L}_{T}$ which has three components: a photometric discrepancy term $\mathcal{L}_{P}$, smoothness term $\mathcal{L}_{S}$ and left-right consistency term $\mathcal{L}_{LR}$, with different weighting parameters $\lambda_{0}, \lambda_{1}, \text{ and } \lambda_{2}$.
\begin{equation}
    \mathcal{L}_{T} =  \lambda_{0} \mathcal{L}_{P} + \lambda_{1} \mathcal{L}_{LR} + \lambda_{2} \mathcal{L}_{S}
    \label{eqn:total}
\end{equation}

Photometric discrepancy term $\mathcal{L}_{P}$ is a sum of an $L1$ loss and a structural dissimilarity term based on SSIM \cite{wang2004image} with multiple window sizes. In our experiments we use $S = \lbrace 3, 5, 7 \rbrace$. $N$ in Equation~\ref{eqn:photo_loss} is the total number of pixels. 

\begin{equation}
\begin{aligned}
    \mathcal{L}_{P} = &\frac{\lambda^{p}_{1}}{N} (\norm{\tilde{\mathcal{X}_{L}}-\mathcal{X}_{L}}_{1} + \norm{\tilde{\mathcal{X}_{R}}-\mathcal{X}_{R}}_{1}) + \\
    & \frac{\lambda^{p}_{s}}{N}\sum_{s\in S}{(SSIM_{s}(\tilde{\mathcal{X}_{L}}, \mathcal{X}_{L}) + SSIM_{s}(\tilde{\mathcal{X}_{R}}, \mathcal{X}_{R}))}\\
\end{aligned}
\label{eqn:photo_loss}
\end{equation}

Depth prediction from photo-consistency is an ill-posed inverse problem, where there are a large number of photo-consistent geometric structures for a given stereo pair. Imposing regularization terms encourages the predictions to be closer to the physically valid solutions. Therefore, we use an edge-aware smoothness regularization term $\mathcal{L_{S}}$ in Equation~\ref{eqn:depthsmootheness} and left-right consistency loss \cite{godard} Equation~\ref{eqn:leftright}. $\mathcal{L}_{LR}$ is computed by warping disparities and comparing them to the original disparities predicted by the network. 

\begin{equation}
    \begin{aligned}
        \label{eqn:leftright}
        & \mathcal{L_{LR}} =  \norm{\tilde{\mathcal{D_{L}}}-\mathcal{{D}_{L}}}_{1} + \norm{\tilde{D_{R}}-\mathcal{D}_{R}}_{1}
        \\
        &\text{where } \tilde{\mathcal{D_{L}}} = \Phi(\mathcal{D}_{R}, \mathcal{D}_{L}) \text{ and } \tilde{\mathcal{D_{R}}} = \Phi(\mathcal{D}_{L}, \mathcal{D}_{R})
        \end{aligned}
\end{equation}

Smoothness term $\mathcal{L_{S}}$ forces disparities $\mathcal{D_{L}} \text{ and } \mathcal{D_{R}}$ to have small gradient magnitudes in smooth image regions. However, the term allows large disparity gradients in regions where there are strong image gradients.

\begin{equation}
\label{eqn:depthsmootheness}
    \centering
    \begin{aligned}
        \mathcal{L_{S}} = &\nabla_{x}{\mathcal{D}_{L}}e^{(-\nabla_{x}{\mathcal{X}_{L}})} + 
    \nabla_{y}{\mathcal{D}_{L}}e^{(-\nabla_{y}{\mathcal{X}_{L}})} +
    \\
    &\nabla_{x}{\mathcal{D}_{R}}e^{(-\nabla_{x}{\mathcal{X}_{R}})} + 
    \nabla_{y}{\mathcal{D}_{R}}e^{(-\nabla_{y}{\mathcal{X}_{R}})}
    \end{aligned}
    \footnotemark
\end{equation}

\footnotetext{$\nabla_{x}$ \text{ is gradient w.r.t x, similarly} $\nabla_{y}$ \text{ is gradient w.r.t y}}

Bilinear sampling, $\Phi$ allows back-propagation of error (sub-)gradients from the output such as $\Tilde{\mathcal{X}}s$ back to the input images $\mathcal{X}$s and disparities $\mathcal{D}$s. Therefore, it is possible to use standard back-propagation to train our network. Formal derivations for the back-propagation of gradients via a bilinear sampling module could be found in~\cite{jaderberg2015spatial}. 

\subsection{Forward mapping}\label{sec:fwdmap}

\begin{figure}[htb!]
\begin{minipage}[b]{1.0\linewidth}
  \centering
 \centerline{\includegraphics[width=\linewidth]{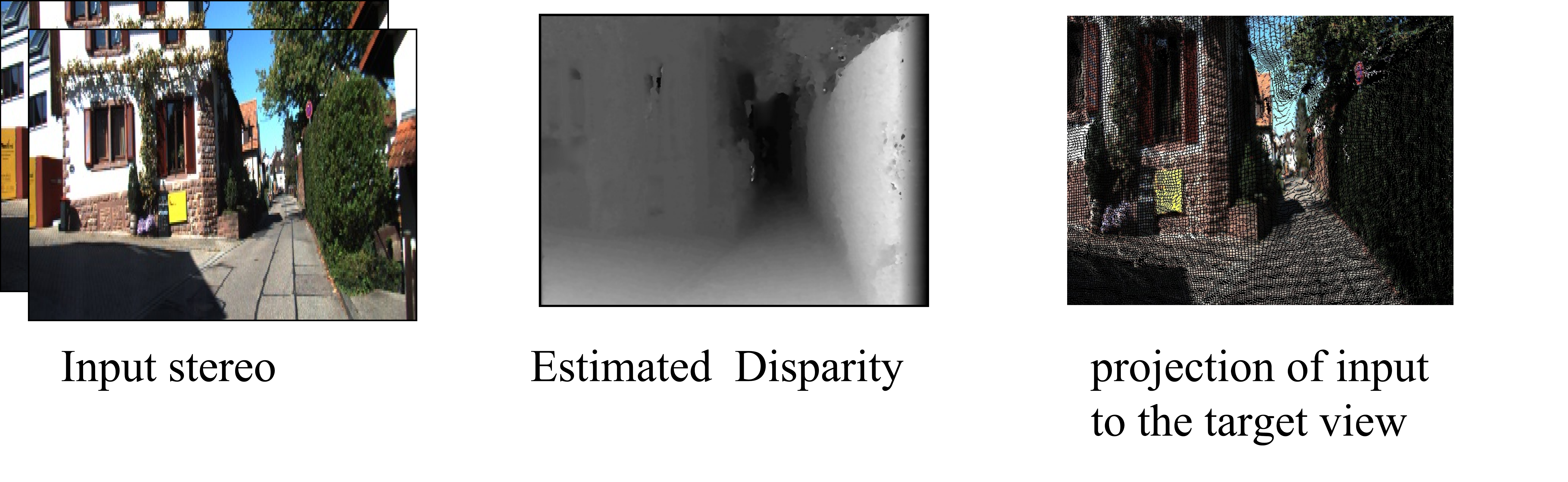}}
  \centerline{(a) Sample view-warping}
  \end{minipage}
\begin{minipage}[b]{1.0\linewidth}
  \centering
  \centerline{\includegraphics[width=\linewidth]{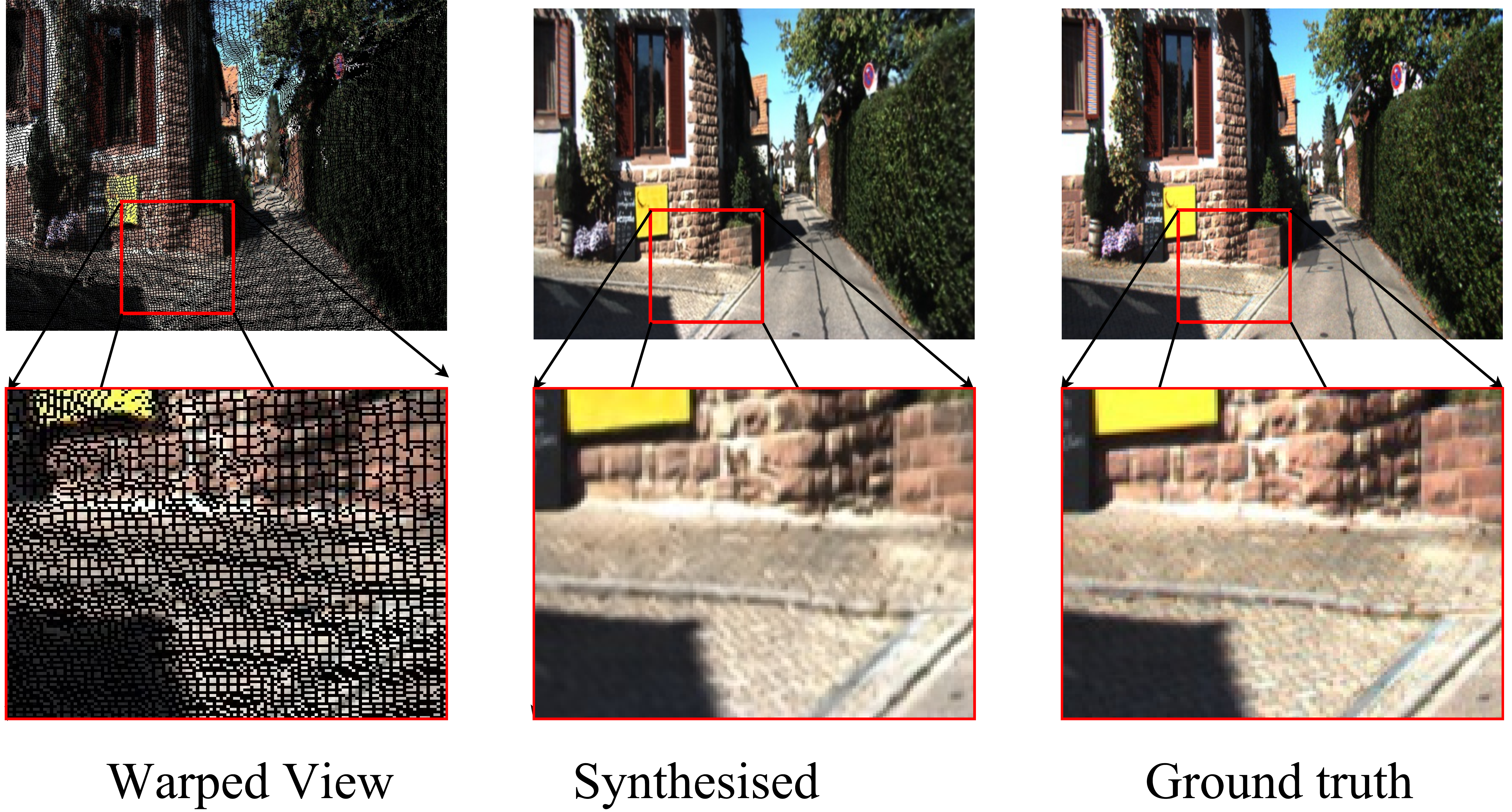}}
  \centerline{(b) Close-up view of warped and target images}
  \end{minipage}
\caption{View-warping with forward mapping.}
\label{fig:warpedview}
\end{figure}

We apply our trained stereo depth predictor on the $\mathcal{V}$ input stereo pairs $\{\{\mathcal{X}^{1}_{L}, X^{1}_{R}\},\{\mathcal{X}^{2}_{L}, \mathcal{X}^{2}_{R}\}, \dots,  \{\mathcal{X}^{V}_{L}, \mathcal{X}^{V}_{R}\}\}$ and generate disparities for the left camera input views, $\{\mathcal{D}^{1}_{L}, \mathcal{D}^{2}_{L}, \dots \mathcal{D}^{V}_{L}\}$. The right stereo views are used only for estimating disparities. Forward mapping and texture inpainting are done only on the images from the left camera. In this section, unless specified otherwise, we refer the images from left camera as the input images/views and we drop the subscripts $L$ and $R$.

Forward-mapping projects input views to the target-view using their respective depth-maps. The predicted disparities of the input views could be converted to depth values. Depth $\mathcal{Z}^{i}_{w,h}$ for a pixel at location $(w,h)$ in the $i-th$ input view, can be computed from the corresponding disparity $\mathcal{D}^{i}_{w,h}$, as follows:
\begin{equation}
    \mathcal{Z}^{i}_{w,h} = \frac{f_{x}*B}{\mathcal{D}^{i}_{w,h}}
\end{equation}
where $K$ is the intrinsic camera matrix, $B$ is the baseline, and $f_{x}$ is focal length.

The goal of forward mapping is to project the input views to the target view, $t$. Given the relative pose between the input-view $i$ and target-view as a transformation matrix $P^{i} = [R^{i}|T^{i}]$, pixel $p_{h,w}^{i}$~(pixel location $\{h,w\}$ in view $i$) will be forward mapped as follows, to a pixel location $p^{t}_{x,y}$ on the target view:

\begin{equation}
\begin{aligned}
&\left[x',y',z' \right] \sim  KP^{i}Z^{i}_{h,w}K^{-1}[h,w,1]^{T} \\
&x = \left \lfloor{x'/z}\right \rfloor \text{ and } y = \left \lfloor{y'/z}\right \rfloor \\
\end{aligned}
\label{eqn:one}
\end{equation}

Following a standard forward projection Equation~\ref{eqn:one}, the reference input frames $\{\mathcal{X}^{1}, \mathcal{X}^{2}, \dots, \mathcal{X}^{V}\} $ are warped to the target view $\{\mathcal{W}^{1}, \mathcal{W}^{2}, \dots, \mathcal{W}^{V}\}$. As shown in Figure~\ref{fig:warpedview}, forward-mapped views have a large number of pixel locations with unknown color values. These holes are created for various reasons. First, forward-mapping is a one-to-one mapping of pixels from the input views to the target view. This creates holes as it doesn't account for zooming in effects of camera movements, which could lead to one-to-many mapping. Moreover, occlusion and rounding effects lead to more holes. In addition to holes, some warped pixels have wrong color values due to inaccuracies in depth prediction.

\subsection{Texture Inpainting}
The goal of our texture inpainting network is to learn generating the target view $\mathcal{X}^{t}$ from the set of warped input views $\{\mathcal{W}^{1}, \mathcal{W}^{2}, \dots, \mathcal{W}^{V}\}$. This is a structured prediction task where the input and output images are aligned. Due to the effects mentioned above in section~\ref{sec:fwdmap}, texture mapping results in noisy warped views, see Figure~\ref{fig:warpedview}. Forward-mapped images $\mathcal{W}^{i}s$ contain two kinds of pixels: \textit{noisy-pixels}, those with unknown (or wrong) color values and \textit{good-pixels}, those with correct color value. Ideally we would like to hallucinate the correct color value for the noisy pixels while maintaining the \textit{good} pixels.

The architecture of our proposed \textit{inpainting} network is inspired by Densely Connected~\cite{huang2017densely} and Residual~\cite{he2016deep} network architectures. Details of the network architecture are presented in Table~\ref{tabel:texture_network}. The network has residual layers with long range skip-connections that feed features from early layers to the top layers. The architecutre design used on our network is designed to facilitate flow of activations as well as gradients through the network. The texture inpainting network is trained by minimizing L1 loss btween the predicted novel views and the original images.

\begin{table}[htb!]
\centering
\begin{tabular}{|l|l|l|l|l|}
\hline
\multicolumn{1}{|c|}{\textbf{Block}} & \textbf{Layer} & \textbf{Input}                   & \textbf{\begin{tabular}[c]{@{}l@{}}Input,Out.\\ channels\end{tabular}} & \textbf{Output size} \\ \hline
Block 0                              & conv\_0        & warped views                     & 4*4,32                                                                 & H, W                 \\ \hline
Block 0                              & res\_1         & conv\_0                          & 32,32                                                                  & H, W                 \\ \hhline{|=|=|=|=|=|}
Block 1                              & res\_2         & pool(res\_1), pool(warped views) & 32+16, 48                                                              & H/2, W/2             \\ \hline
Block 1                              & res\_3         & res\_2                           & 48 ,48                                                                 & H/2, W/2             \\ \hline
Block 1                              & res\_4         & res\_3                           & 48 ,48                                                                 & H/2, W/2             \\ \hline
Block 1                              & res\_5         & res\_4                           & 48 ,48                                                                 & H/2, W/2             \\ \hline
Block 1                              & res\_6         & res\_5                           & 48 ,48                                                                 & H/2, W/2             \\ \hline
Block 1                              & res\_7         & res\_6                           & 48 ,48                                                                 & H/2, W/2             \\ \hline
Block 1                              & res\_8         & res\_7                           & 48 ,48                                                                 & H/2, W/2             \\ \hhline{|=|=|=|=|=|}
Block 2                              & conv\_9        & upsample(res\_8),  res\_1        & 48+32,48                                                               & H, W                 \\ \hline
Block 2                              & res\_10        & conv\_9                          & 48, 48                                                                 & H, W                 \\ \hhline{|=|=|=|=|=|}
Block 2                              & res\_11        & res\_10                          & 48, 48                                                                 & H,W                  \\ \hline
Block 3                              & conv\_12       & res\_11                          & 48, 16                                                                 & H,W                  \\ \hline
Block 3                              & output         & conv\_12                         & 16, 3                                                                  & H,W                  \\ \hline
\end{tabular}

\caption{\textbf{Texture inpainting network architecture.} The network is mainly residual, except special convolution layers which could be used as input and output layers. Similar to \textit{Densely Connected Nets} \cite{huang2017densely}, at the beginning of each block, the number of channels in the concatenated feature maps could be decreased via a convolutional layer.}
\label{tabel:texture_network}
\end{table}
\squeezeup

\begin{figure}[htb!]
  \centering
  \begin{subfigure}{\linewidth}
    \centering
    \includegraphics[height=6cm,width=\linewidth]{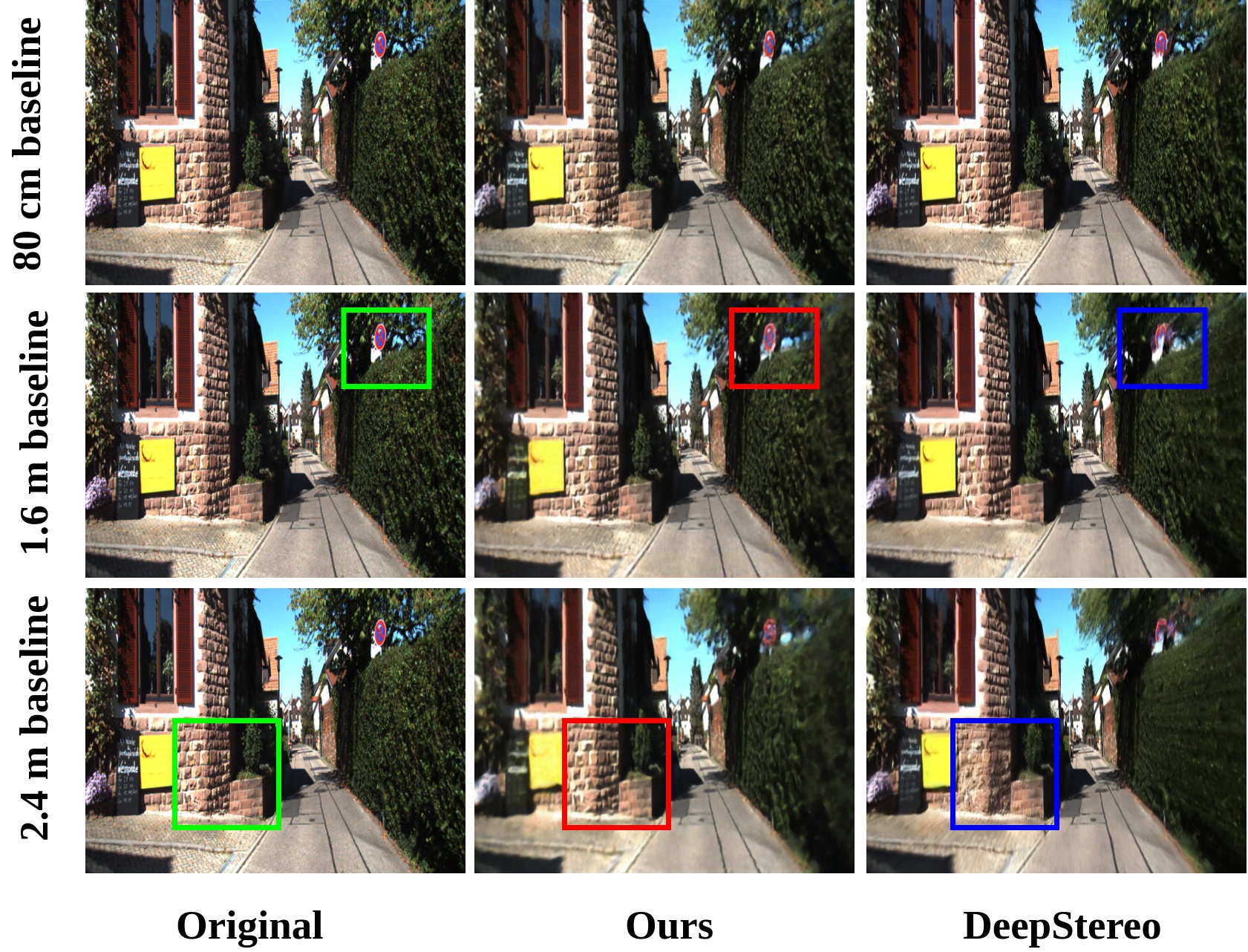}
    \caption{Rendering with our approach and DeepStereo \cite{flynn2015deepstereo} }
  \end{subfigure}
\vspace{0.5cm}
  \begin{subfigure}{\linewidth}
    \centering
    \includegraphics[height=4cm,width=0.9\linewidth]{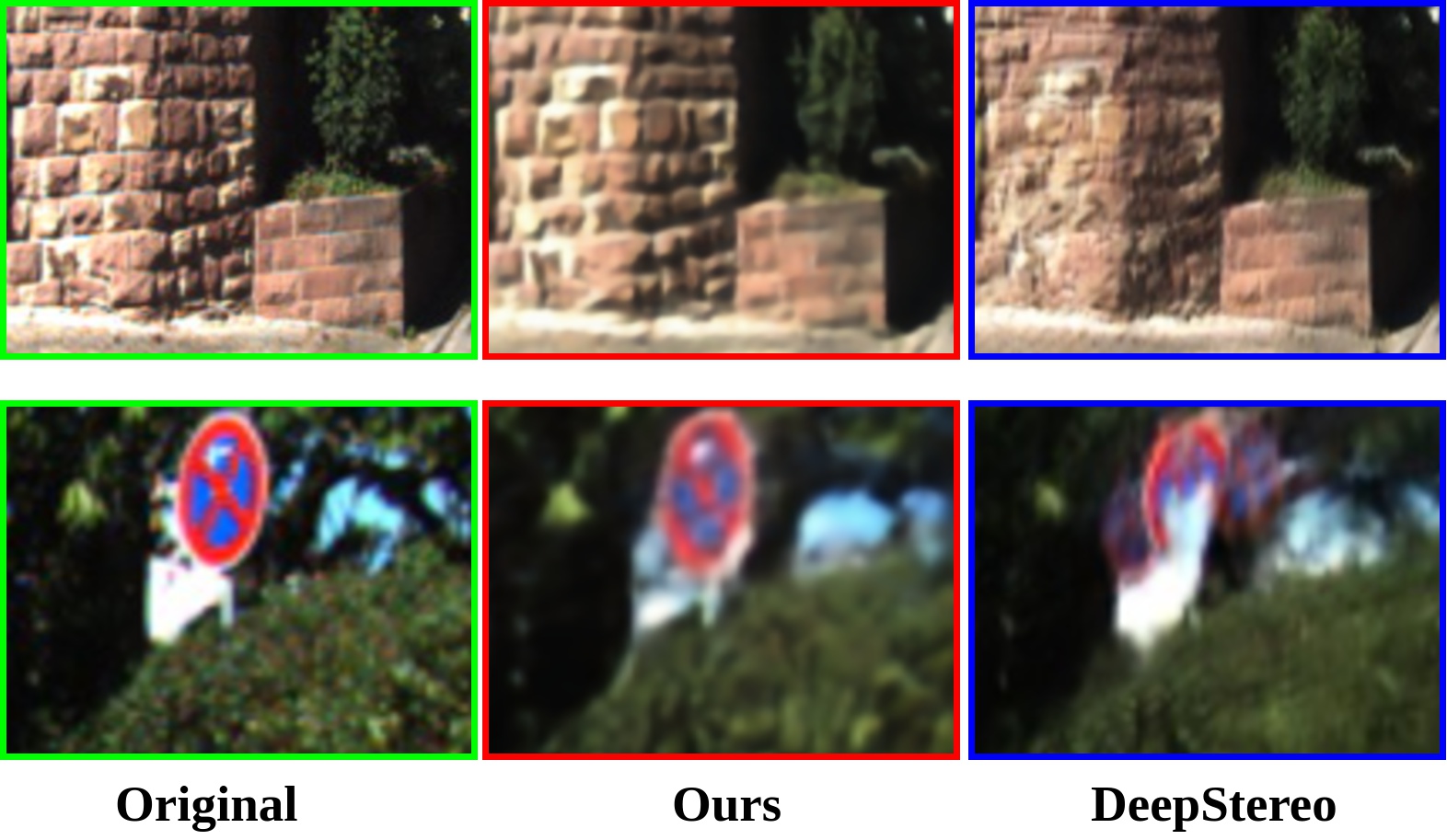}
    \caption{Close-up views.}
  \end{subfigure}
  \caption{Rendering of a sample scene for qualitative evaluation. Close-up views show that preserves the geometric structure better than \textit{DeepStereo} and textit{DeepStereo} has ghosting where the traffic sign is replicated multiple times. Our rendering resembles the target except for slight amount of blur.}
  \label{fig:qualitative}
\end{figure}
\squeezeup

\section{Experiments}
We tested our proposed approach on the KITTI \cite{geiger2012we} public computer vision benchmark. Our approach has lower rendering error than the current state-of-the-art~\cite{flynn2015deepstereo}~(see Table~\ref{table:mainresult}). Qualitative evaluation also show that out method better preserves the geometric details of the scene, shown in Figure~\ref{fig:qualitative}. 
\subsubsection*{Network Training and Hyper-parameters}
The depth predictor loss term has three weighting parameters: $\lambda_{0}$ for photometric loss, $\lambda_{1}$ for left-right consistency loss and $\lambda_{2}$ for local smoothness regularization. The weighting parameters have to be set properly, wrong weights might lead trivial solutions such as a constant disparity output for every pixel. The weighting parameters used in our experiments are the following: $\lambda_{0}=5$, $\lambda_{1}=0.01$, $\lambda_{2}=0.0005$. The photometric loss is also a weighted combination of l1 loss and SSIM losses with different window sizes. These weighting factors are set to be $\lambda^{1}_{P}=0.2$, $\lambda^{3}_{P}=0.8$, $\lambda^{5}_{P}=0.2$ and $\lambda^{7}_{P}=0.2$. As discussed in section 3. the depth-prediction and texture inpainting networks are trained separately. We train the networks with back-propagation using Adam optimizer \cite{kingma2014adam}, with mini-batch size 1 and learning rate of 0.0004. The depth prediction network is trained for 200,000 iterations and the inpainting network is trained for 1 Million iterations.
\subsubsection*{Dataset:} We evaluate our proposed method on the publicly available KITTI Odometry Dataset \cite{geiger2012we}. KITTI is an interesting dataset for large-baseline novel view synthesis. The dataset has a large number of frames recorded in outdoors sunny urban environment. The dataset contains variations in the scene structure~(vegetation, buildings, etc), strong illumination changes and noise which makes KITTI a challenging benchmark. The dataset contains around $23000$ stereo pairs divided into 11 sequences~(\textit{00} to \textit{10}). Each sequence contains a set of images captured by a stereo camera, mounted on top of car driving through a city. The stereo camera captures frames every $\approx 80~cms$. Each sequence corresponds to a single drive and extrinsic camera calibrations are available for each sequence separately.

In order to make our results comparable to \textit{DeepStereo}~\cite{flynn2015deepstereo}, we hold out \textit{sequence 04} for validation, \textit{sequence 10} for test and used the rest for training. In our experiments, $5$ consecutive images are used, with the middle one is held out as target and the other $4$ images are used as reference. Tests are done at different values of spacing between consecutive images. Taking consecutive frames gives spacing of around $ 0.8~m$. Sampling every other frame give a spacing of $\approx1.6 m$ and every third frame gives $\approx2.4 m$ spacing. The error metric used to evaluate performance of rendering methods a mean absolute brightness error, computed per pixel per color channel. 
\begin{table}[htb!]
\centering
\begin{tabular}{|l|l|l|l|l|}
\hline
Spacing & Method                  & Test 0.8 m       & Test 1.6 m      & Test 2.4 m       \\ \hline
\multirow{2}{*}{Train 0.8 m }  & Ours       & \textbf{6.66} & \textbf{8.90} & \textbf{12.14}  \\ \cline{2-5} 
                           & DeepStereo & 7.49          & 10.41         & 13.44          \\ \hline
\multirow{2}{*}{Train 1.6 m} & Ours       & \textbf{6.92} & \textbf{8.47} & \textbf{10.38} \\ \cline{2-5} 
                           & DeepStereo & 7.60           & 10.28         & 12.97          \\ \hline
\multirow{2}{*}{Train 2.4 m} & Ours       & \textbf{7.47} & \textbf{8.73} & \textbf{10.28} \\ \cline{2-5} 
                           & DeepStereo & 8.06          & 10.73         & 13.37          \\ \hline
\end{tabular}
\caption{\textbf{Quantitative evaluation of our method against \textit{DeepStereo}}. We used our texture inpainting network with residual blocks as shown in Table~\ref{tabel:texture_network}. Our approach outperforms \textit{DeepStereo}~\cite{flynn2015deepstereo} in all cases. Each row shows the performance of a method that is trained on a specific input camera spacing and tested on all three spacings.}
\label{table:mainresult}

\end{table}
\squeezeup
\squeezeup

\subsubsection*{Results:} Table~\ref{table:mainresult} shows the results of our proposed approach compared to DeepStereo~\cite{flynn2015deepstereo}, on the KITTI dataset. Our approach outperforms \textit{DeepStereo} in all spacings of the input views, while being significantly faster. In Figure~\ref{fig:qualitative}, sample renderings are shown for a qualitative evaluation of our approach and \textit{DeepStereo}. As shown in the lowest two rows of Figure~\ref{fig:qualitative}, renderings of \textit{DeepStereo} suffer from ghosting effects due to cross-talk between different depth planes, which led to the replicated traffic sign and noisy bricks(as could be seen in the close-up views).

We performed tests to compare our texture inpainting architecture against the so called UNet architecture~\cite{ronneberger2015u}. Furthermore, in order to evaluate the significance of the texture inpainting stage, we measured the accuracy of our system when the texture inpainting stage is replaced by simple median filtering scheme applied on the warped views. Table~\ref{table:ablation} shows that our inpainting network (with residual layers) achieves better performance than UNet and the "basic" median filtering baseline. A test performed to investigate the effect of using residual layers instead of convolutional layers shows that residual layers give better performance.


%

\begin{table}[htb!]
\centering
\begin{tabular}{|l|l|l|l|}
\hline
\multicolumn{1}{|c|}{Method}                                        & {\color[HTML]{000000} Test 0.8 m} & {\color[HTML]{000000} Test 1.6 m} & {\color[HTML]{000000} Test 2.4 m} \\ \hline
\begin{tabular}[c]{@{}l@{}}Ours with\\ Residual blocks\end{tabular} & \textbf{6.66}                     & \textbf{8.47}                     & \textbf{10.28}                    \\ \hline
\begin{tabular}[c]{@{}l@{}}Ours with\\  Conv blocks\end{tabular}    & 6.70                              & 8.57                              & 10.34                             \\ \hline
UNet                                                                & 7.08                              & 9.60                              & 11.17                             \\ \hline
\begin{tabular}[c]{@{}l@{}}Median of \\ Warped-views\end{tabular}   & 18.97                             & 31.72                             & 41.29                             \\ \hline
\end{tabular}
\caption{\textbf{Testing different choices of texture inpainting networks.} We test the effect of using residual and convolutional layers. We also compare the performance of our network against the UNet~\cite{ronneberger2015u} architecture and the median of warped input views. In all baselines, our network with residual layers gives the lowest error.}
\label{table:ablation}

\end{table}
\squeezeup

\subsubsection*{Timing}
During the test phase, at a resolution of $528\times384$, our depth prediction stage and the forward mapping take $6.08$ seconds and $2.60$ seconds (for four input neighbouring views), respectively. The texture rendering takes takes $0.05$ seconds. Thus, the total time adds up to $8.73$ seconds, per frame. This is much faster than \textit{DeepStereo} which takes $12$ minutes to render a single frame of resolution $500 \times 500$. Our experiments are performed on a multi-core cpu machine with single Tesla V100 GPU.

\section{Conclusion and Future Work}

We presented a fast and accurate novel view synthesis pipeline based on stereo-vision and convolutional networks. Our method decomposes novel view synthesis into, view-dependent geometry estimation and texture inpainting problems. Thus, our method utilizes the power of convolutional neural networks in learning structured prediction tasks. 

Our proposed method is tested on a challenging benchmark, where most existing approaches are not able to produce reasonable results. The proposed approach is significantly faster and more accurate than the current state-of-the-art. As part of a future work, we would like to explore faster architectures to achieve real-time performance. 




\bibliographystyle{splncs}
\bibliography{egbib.bib}
\end{document}